\documentclass[
]{ceurart}

\usepackage{listings}
\usepackage{hyperref}
\begin{document}

\copyrightyear{2025}
\copyrightclause{Copyright for this paper by its authors.
  Use permitted under Creative Commons License Attribution 4.0
  International (CC BY 4.0).}

\conference{CLEF 2025 Working Notes, September 9 -- 12 September 2025, Madrid, Spain}

\title{QU-NLP at CheckThat! 2025: Multilingual Subjectivity in News Articles Detection using Feature-Augmented Transformer Models with Sequential Cross-Lingual Fine-Tuning}



\author{Mohammad AL-Smadi}[%
orcid=0000-0002-7808-6962,
email=malsmadi@qu.edu.qa,
]
\address{Qatar University,
  Doha, Qatar}




\begin{abstract}
  This paper presents our approach to the CheckThat! 2025 Task 1 on subjectivity detection, where systems are challenged to distinguish whether a sentence from a news article expresses the subjective view of the author or presents an objective view on the covered topic. We propose a feature-augmented transformer architecture that combines contextual embeddings from pre-trained language models with statistical and linguistic features. Our system leveraged pre-trained transformers with additional lexical features: for Arabic we used AraELECTRA augmented with part-of-speech (POS) tags and TF-IDF features, while for the other languages we fine-tuned a cross-lingual DeBERTa~V3 model combined with TF-IDF features through a gating mechanism. We evaluated our system in monolingual, multilingual, and zero-shot settings across multiple languages including English, Arabic, German, Italian, and several unseen languages. The results demonstrate the effectiveness of our approach, achieving competitive performance across different languages with notable success in the monolingual setting for English (rank 1\textsuperscript{st} with macro-F1=0.8052), German (rank 3\textsuperscript{rd} with macro-F1=0.8013), Arabic (rank 4\textsuperscript{th} with macro-F1=0.5771),  and Romanian (rank 1\textsuperscript{st} with macro-F1=0.8126) in the zero-shot setting. We also conducted an ablation analysis that demonstrated the importance of combining TF-IDF features with the gating mechanism  and the cross-lingual transfer for subjectivity detection.  Furthermore, our analysis reveals the model's sensitivity to both the order of cross-lingual fine-tuning and the linguistic proximity of the training languages.
\end{abstract}

\begin{keywords}
  Subjectivity Detection \sep
  Multilingual NLP \sep
  Cross-lingual Transfer \sep
  Transformer Models \sep
  AraELECTRA \sep
  DeBERTa V3 \sep
  TF-IDF Features \sep
  POS Tagging \sep
  Zero-shot Learning 
\end{keywords}

\maketitle

\section{Introduction}

The rapid increase in online news and social media posts has led to an crucial need for automated tools that can distinguish between factual reporting and opinion-based content. Subjectivity detection is defined as the task of identifying whether a text expresses personal opinions, beliefs, feelings, or judgments versus presenting only factual information \cite{ruggeri2023definition}. Subjectivity detection has become a critical component in various natural language processing applications, including media bias detection, stance detection, and fact-checking services. Moreover, Subjectivity detection tools have the ability to automatically identify subjective content in multilingual contexts, whereas manual analysis is expensive and time consuming across different languages.

The CheckThat! Lab at CLEF 2025 \cite{clef-checkthat:2025:task1} introduced Task 1 on subjectivity detection, challenging participants to develop systems capable of classifying sentences from news articles as either subjective (SUBJ) or objective (OBJ). This task was structured into three distinct settings: monolingual (training and testing in the same language), multilingual (training and testing on data comprising several languages), and zero-shot (training on several languages and testing on unseen languages). This comprehensive evaluation framework allows for a thorough assessment of systems' capabilities to generalize across languages and domains.

In this paper, we present the approach developed by our team QU-NLP\footnote{the team name was set to the username default value of "msmadi" on the codalab website, See tasks' final results on \url{https://gitlab.com/checkthat_lab/clef2025-checkthat-lab/-/tree/main/task1}} for the CheckThat! 2025 Task 1. Our models leverage a feature-augmented transformer architecture that combines the contextual learning capabilities of pre-trained language models with statistical and linguistic features specifically selected to capture signs of subjectivity. As the task covers different language settings, we employed tailored models: (a) a specialized AraELECTRA-based model for Arabic and (b) a DeBERTa-based architecture with sequential cross-lingual fine-tuning for other languages.

Our contributions can be summarized as follows:
\begin{itemize}
    \item We propose a feature-augmented transformer architecture that effectively combines deep contextual representations with explicit linguistic features for subjectivity detection.
    \item We demonstrate the effectiveness of sequential cross-lingual fine-tuning for improving performance in multilingual and zero-shot settings.
    \item We provide a comprehensive analysis of our system's performance across different languages and settings, highlighting strengths and limitations.
    \item We investigate the contribution of different feature combinations to the overall performance, offering insights into the linguistic markers of subjectivity across languages.
\end{itemize}

The remainder of this paper is organized as follows: Section 2 reviews related work in subjectivity detection and multilingual NLP. Section 3 describes the task, datasets, and our methodology, including model architecture and training setup. Section 4 presents our experimental results across different languages and settings. Section 5 discusses our findings, analyzes error cases, and explores the implications of our results. Finally, Section 6 concludes the paper and suggests directions for future work.

\section{Related Work}

Subjectivity detection has been an active area of research in natural language processing for over two decades. Early approaches to this task relied heavily on lexical resources and hand-crafted features \cite{wiebe2004learning}, while more recent methods leverage deep learning architectures and transfer learning from pre-trained language models. In this section, we review relevant literature on subjectivity detection, multilingual approaches to text classification, and recent advances in cross-lingual transfer learning.

\subsection{Multilingual Text Classification}

Multilingual text classification has gained significant attention with the development of cross-lingual embeddings and multilingual pre-trained language models. Cross-lingual transfer learning aims to leverage knowledge from resource-rich languages to improve performance on low-resource languages. Various approaches have been proposed to enhance cross-lingual transfer, including adversarial training, meta-learning, and language-specific adapters. Artetxe and Schwenk \cite{artetxe2019massively} proposed a language-agnostic sentence embedding model trained on parallel data from 93 languages, enabling zero-shot cross-lingual transfer for various classification tasks.

Multilingual pre-trained language models such as mBERT \cite{devlin2019bert}, XLM-R \cite{conneau2020unsupervised}, and mT5 \cite{xue2021mt5} have become the foundation for state-of-the-art multilingual text classification systems. These models are pre-trained on massive multilingual corpora, allowing them to learn shared representations across languages that can be fine-tuned for specific downstream tasks.

Several studies have explored techniques to improve cross-lingual transfer in text classification. Wu and Dredze \cite{wu2019beto} analyzed the cross-lingual capabilities of mBERT across 39 languages and 5 NLP tasks, finding that it performs remarkably well even for languages with limited pre-training data. Pires et al. \cite{pires2019multilingual} investigated the structural similarities captured by mBERT that enable its cross-lingual abilities, showing that it aligns representations of similar words across languages.

Yan et al. \cite{chen2019multi} proposed a meta-learning approach for cross-lingual transfer, where a model learns to quickly adapt to new languages with minimal supervision. Pfeiffer et al. \cite{pfeiffer2020mad} introduced MAD-X, a modular adaptation framework that uses language adapters to enable parameter-efficient cross-lingual transfer.

Sequential fine-tuning has emerged as an effective technique for cross-lingual transfer. Do and Gaspers  \cite{tran2020cross} demonstrated that sequentially fine-tuning a multilingual model on related languages before the target language can significantly improve performance. Similarly, Nooralahzadeh et al. \cite{lauscher2020zero} showed that intermediate fine-tuning on a related high-resource language can boost zero-shot performance on low-resource languages.

\subsection{Subjectivity Detection in News Media}

Subjectivity detection in news media presents unique challenges due to the different ways in which subjective content can be expressed in seemingly objective reporting. The task of distinguishing between subjective and objective text has its roots in the pioneering work of Wiebe et al. \cite{wiebe1999development}, who created one of the first annotated corpora for subjectivity analysis. This early work established the foundation for subsequent research on subjectivity detection, sentiment analysis, and opinion mining.  Recasens et al. \cite{recasens2013linguistic} identified linguistic indicators of bias in news articles, including factive verbs, implicative verbs, and hedges, which can signal subjective content without explicit opinion markers.

Subjectivity detection in Arabic news has gained increasing attention over the past two decades, with researchers aiming to distinguish between factual reporting and opinionated content in Arabic-language media. Early foundational work by El-Halees~\cite{elhalees2011} explored text classification in Arabic news using machine learning techniques such as maximum entropy, laying the groundwork for subsequent efforts in identifying subjective language in formal Arabic contexts. Abdul-Mageed and Diab~\cite{abdulmageed2011} advanced the field by developing supervised models to detect subjectivity and sentiment in Modern Standard Arabic, demonstrating the viability of using linguistic features and annotated corpora for reliable classification. More recent research by Al-Smadi et al.~\cite{alsmadi2016} introduced an aspect-based sentiment analysis framework tailored to Arabic news articles, marking a shift from document-level to aspect-level opinion mining. While not the first to explore subjectivity in Arabic, this study is notable for its emphasis on identifying sentiment tied to specific news aspects, thereby offering a more nuanced understanding of reader affect. 

Recent work has focused on developing fine-grained approaches to detect different types of subjectivity in news. Spinde et al. \cite{spinde2021automated} created a comprehensive framework for detecting media bias, incorporating subjectivity detection as a key component.

The CheckThat! Lab has contributed significantly to advancing research in this area by providing multilingual benchmarks for subjectivity detection in news. The annotation guidelines developed by Ruggeri et al. \cite{ruggeri2023definition} provide a language-agnostic framework for identifying subjectivity, enabling consistent annotation across different languages. Building on this work, Antici et al. \cite{antici2024corpus} created a corpus for sentence-level subjectivity detection in English news articles, while Suwaileh et al. \cite{suwaileh2024thatiar} developed ThatiAR, a dataset for subjectivity detection in Arabic news sentences.

Our work builds upon these foundations, leveraging insights from both subjectivity detection research and cross-lingual transfer learning to develop a robust system for multilingual subjectivity detection in news media.

\section{Research Methodology}

\subsection{Task Description}

The CheckThat! 2025 Task 1 focused on subjectivity in news articles detection. Participants were requested to develop systems capable of distinguishing whether a sentence from a news article expresses the subjective view of the author or presents an objective view on the covered topic. This binary classification task required systems to label text sequences as either subjective (SUBJ) or objective (OBJ).

The task was structured into three distinct evaluation settings:

\begin{enumerate}
    \item \textbf{Monolingual}: Systems were trained and tested on data in a single language. This setting was implemented for five languages: English, Arabic, Italian, and German.
    
    \item \textbf{Multilingual}: Systems were trained and tested on data from several languages. 
    
    \item \textbf{Zero-shot}: Systems were trained on several languages from the settings above and tested on unseen languages (mainly Polish, Ukrainian, Romanian, and Greek).
\end{enumerate}

The participating systems were ranked based on their macro-averaged F1 score, which equally weights the performance on both the SUBJ and OBJ classes. 

\subsection{Dataset}

The dataset provided for the task consisted of sentences extracted from news articles in multiple languages, manually annotated as either subjective (SUBJ) or objective (OBJ). For each language, the data was divided into three sets: training, development, and test.

Table \ref{tab:dataset_stats} presents the statistics of the dataset for each language. The data exhibits some class imbalance, with objective sentences generally outnumbering subjective ones across most languages. This imbalance varies across languages, with Arabic having the largest dataset (3,661 annotated sentences) and German having the smallest (1,628 annotated sentences). About 300 sentences were provided as test dataset for each language.

\begin{table}
  \caption{Dataset Statistics for CheckThat! Task 1 - Subjectivity}
  \label{tab:dataset_stats}
  \begin{tabular}{lcccccc}
    \toprule
    \multirow{2}{*}{Language} & \multicolumn{2}{c}{Train} & \multicolumn{2}{c}{Dev} & \multicolumn{2}{c}{Dev-Test} \\
    \cmidrule(r){2-3} \cmidrule(r){4-5} \cmidrule(r){6-7}
    & OBJ & SUBJ & OBJ & SUBJ & OBJ & SUBJ \\
    \midrule
    English & 532 & 298 & 222 & 240 & 362 & 122 \\
    Italian & 1231 & 382 & 490 & 177 & 377 & 136 \\
    German & 492 & 308 & 317 & 174 & 226 & 111 \\
    Arabic & 1391 & 1055 & 266 & 201 & 425 & 323 \\
    \bottomrule
  \end{tabular}
\end{table}

The annotation of the dataset followed the guidelines developed by Ruggeri et al. \cite{ruggeri2023definition}, which provide a language-agnostic framework for identifying subjectivity in news text. These guidelines define subjective content as text that expresses personal opinions, beliefs, or judgments, while objective content presents factual information without expressing the author's perspective. The reader is redirected to \cite{antici2024corpus, suwaileh2024thatiar, clef-checkthat:2025:task1} for more information about the datasets.

\subsection{Models}

Our approach to the subjectivity detection task involved developing two distinct model architectures tailored to different language settings. For the Arabic monolingual task, we designed a specialized model leveraging AraELECTRA with additional linguistic features. For all other settings (monolingual non-Arabic, multilingual, and zero-shot), we employed a DeBERTa-based architecture with sequential cross-lingual fine-tuning. The upcoming sub-sections explain in more detail the models architectures along with thier training setups.

\subsubsection{Arabic Monolingual Model}

We developed a feature-augmented transformer architecture for Arabic, leveraging the AraELECTRA model \citep{antoun2020araelectra}. This architecture integrates the pre-trained language model's contextual understanding with supplementary linguistic features. Specifically, it incorporates Part-of-Speech (POS) tags and Term Frequency-Inverse Document Frequency (TF-IDF) representations to capture subjectivity markers in Arabic text.

The proposed model builds upon ELECTRA \citep{clark2020electra} and its Arabic adaptation, AraELECTRA \citep{antoun2020araelectra}. ELECTRA is an encoder-only transformer designed for enhanced efficiency in Natural Language Processing (NLP) tasks. Unlike traditional Masked Language Models (MLMs), ELECTRA employs a "replaced token detection" training strategy. While models like BERT \citep{devlin2018bert} predict masked words, ELECTRA's generator component proposes plausible alternative tokens. A discriminator then identifies whether each input token is original or replaced. This unique strategy compels the model to learn from all input tokens, rather than just masked ones. Consequently, this approach boosts model efficiency and reduces the required training epochs.

The model consists of the following components:

\begin{enumerate}
    \item \textbf{Backbone Encoder}: We used the pre-trained \texttt{araelectra-base-discriminator}\footnote{\url{https://huggingface.co/aubmindlab/araelectra-base-discriminator}} as the core of our model \citep{antoun2020araelectra}. The [CLS] token from the final hidden layer is passed through a self-attention module (MultiheadAttention) to obtain a refined representation.
    
    \item \textbf{Part-of-Speech Features}: We extracted POS tag distributions using the \texttt{bert-base-arabic-camelbert-mix-pos-msa}\footnote{\url{https://huggingface.co/CAMeL-Lab/bert-base-arabic-camelbert-msa}} model \cite{inoue-etal-2021-interplay}. The resulting 9-dimensional POS tag distribution is projected to 64 dimensions via a linear layer followed by Rectified Linear Unit (ReLU) activation function. Applied after linear layers, ReLU enables models to learn complex data patterns \citep{glorot2011deep}. This boosts the model's ability to recognize deep and complex relationships.
    
    \item \textbf{TF-IDF Features}: We computed TF-IDF features over character n-grams (3-7) using a TfidfVectorizer. The resulting vector is reduced to 128 dimensions through a learnable projection layer with ReLU activation.
    
    \item \textbf{Fusion and Classification}: The refined [CLS] embedding from AraELECTRA (768 dimensions), the POS projection (64 dimensions), and the TF-IDF projection (128 dimensions) are concatenated into a 960-dimensional feature vector. This vector is then passed through a fully connected network consisting of a linear layer (960 → 512) followed by LayerNorm and Dropout, and a final linear layer (512 → 2) for binary classification.
\end{enumerate}

\subsubsection{DeBERTa-based Model for Other Languages}

For non-Arabic languages and the multilingual/zero-shot settings, we developed a model based on the DeBERTa V3 architecture \citep{he2023debertav3improvingdebertausing} with a gating mechanism for integrating lexical features. This model was designed to effectively transfer knowledge across languages through sequential fine-tuning.

The model architecture includes:

\begin{enumerate}
    \item \textbf{DeBERTa V3 Encoder}: We used the \texttt{deberta-v3-large}\footnote{\url{https://huggingface.co/microsoft/deberta-v3-large}} model as our backbone. The encoder outputs are passed through a 16-head self-attention layer to capture richer inter-token dependencies. We extract the representation corresponding to the [CLS] token and apply layer normalization and dropout.
    
    \item \textbf{TF-IDF Lexical Branch}: We extract lexical features using a TfidfVectorizer with character n-grams (3-7). The resulting sparse matrix is projected into a dense 128-dimensional vector via a feedforward layer.
    
    \item \textbf{Gating Mechanism}: A gating scalar is computed to dynamically weigh the importance of lexical versus contextual information. This gate modulates the 128-dimensional TF-IDF embedding.
    
    \item \textbf{Feature Fusion and Classification}: The gated TF-IDF vector and the DeBERTa-derived CLS embedding are concatenated and passed through a classification head consisting of linear layers, layer normalization, ReLU activation, and dropout.
\end{enumerate}
\subsection{Gating Mechanism for Feature Fusion}

To effectively integrate sparse lexical representations with dense contextual embeddings, our model employs a learnable \textit{gating mechanism} that dynamically modulates the contribution of TF-IDF features based on the semantic richness of the input as captured by DeBERTaV3.

Let $\mathbf{h}_{\text{BERT}} \in \mathbb{R}^{d}$ denote the contextualized representation derived from the [CLS] token output of the DeBERTaV3 encoder, where $d$ is the hidden size of the transformer. The TF-IDF vector, denoted $\mathbf{h}_{\text{TFIDF}} \in \mathbb{R}^{k}$, is passed through a fully connected layer with ReLU activation to yield $\tilde{\mathbf{h}}_{\text{TFIDF}} \in \mathbb{R}^{128}$, enhancing its representational capacity. The gating mechanism then computes a scalar gate value:

\[
g = \sigma(\mathbf{W}_g \mathbf{h}_{\text{BERT}} + b_g)
\]

where $\mathbf{W}_g \in \mathbb{R}^{1 \times d}$, $b_g \in \mathbb{R}$, and $\sigma(\cdot)$ is the sigmoid activation function. This scalar gate $g \in [0,1]$ acts as a dynamic weighting coefficient:

\[
\hat{\mathbf{h}}_{\text{TFIDF}} = g \cdot \tilde{\mathbf{h}}_{\text{TFIDF}}
\]

The modulated TF-IDF vector $\hat{\mathbf{h}}_{\text{TFIDF}}$ is then concatenated with $\mathbf{h}_{\text{BERT}}$ to form the joint representation:

\[
\mathbf{h}_{\text{joint}} = [\mathbf{h}_{\text{BERT}}; \hat{\mathbf{h}}_{\text{TFIDF}}]
\]

This joint vector is subsequently passed through a feedforward layer followed by a classifier to produce the final output logits.

The gating mechanism enables the model to adaptively regulate the influence of TF-IDF features on a per-instance basis. When semantic signals from the pretrained language model are strong, the gate may downscale the TF-IDF contribution. Conversely, in scenarios where domain-specific vocabulary or sparse lexical cues offer additional value, the gate enhances their impact. This dynamic fusion strategy improves robustness across domains and languages by learning to balance deep semantic understanding with interpretable lexical signals. This approach draws inspiration from prior work on \textit{Highway Networks}~\citep{srivastava2015highwaynetworks} and feature gating mechanisms in multimodal learning, where learned gates enable networks to dynamically fuse heterogeneous input modalities.

\subsection{Training Setup}

\subsubsection{Arabic Monolingual Model Training}

For the Arabic model, we employed the following training configuration:

\begin{itemize}
    \item \textbf{Preprocessing}: Input text was tokenized using the ELECTRA tokenizer with a maximum length of 512 tokens. POS tag distributions were normalized, and TF-IDF vectors were computed with a maximum of 3000 features and a minimum document frequency of 2.
    
    \item \textbf{Training Parameters}: We used a learning rate of 1e-5, a batch size of 16, and gradient accumulation of 4 steps. The model was trained for up to 100 epochs with early stopping (patience = 3) based on evaluation loss. We applied weight decay of 0.01 and enabled mixed precision training (fp16) for efficiency.
    
    \item \textbf{Optimization}: We used the AdamW optimizer with a linear learning rate scheduler and 100 warmup steps.
    
    \item \textbf{Evaluation}: The model was evaluated after each epoch using the development set, and the best checkpoint was selected based on the lowest evaluation loss.
\end{itemize}

\subsubsection{DeBERTa-based Model Training}

For the DeBERTa-based model, we implemented a sequential cross-lingual fine-tuning approach:

\begin{itemize}
    \item \textbf{Preprocessing}: Sentences were tokenized using the DeBERTaV2Tokenizer with a maximum length of 512 tokens. TF-IDF features were extracted from the training data and saved for later use.
    
    \item \textbf{Sequential Fine-tuning}: We trained the model in a specific language sequence: [German → Italian → English]. Starting with the base \texttt{microsoft/deberta-v3-large} checkpoint, we fine-tuned on German data, then used the resulting model to fine-tune on Italian data, and finally fine-tuned on English data.
    
    \item \textbf{Training Parameters}: We used a learning rate of 1e-5, a batch size of 8, and gradient accumulation of 2 steps. Each language-specific fine-tuning was run for up to 100 epochs with early stopping (patience = 2) based on evaluation loss. We applied weight decay of 0.01 and used a cosine learning rate scheduler with 100 warmup steps.

    \item \textbf{Multilingual and Zero-shot Setting}: For both multilingual and zero-shot evaluation, we evaluated the model fine-tuned on the sequence of languages (German → Italian → English) without any additional training on the target languages.
\end{itemize}

\section{Results}

In this section, we present the results of our systems across the three evaluation settings: monolingual, multilingual, and zero-shot. We compare our performance with other participating teams and analyze the effectiveness of our approaches for different languages. For more information about the baseline models or the other participating teams' models the reader is redirected to \cite{clef-checkthat:2025:task1}.

\subsection{Monolingual Results}

Table \ref{tab:monolingual_results} presents the results of our system in the monolingual setting for each of the four languages, along with the performance of the top-performing team and the baseline for each language.

\begin{table}
  \caption{Monolingual Results (Macro F1 scores)}
  \label{tab:monolingual_results}
  \begin{tabular}{lccc}
    \toprule
    Language & QU-NLP (Our Team) & Top Team & Baseline \\
    \midrule
    English & 0.8052 (1st) & 0.8052 (QU-NLP) & 0.5370 \\
    German & 0.8013 (3rd) & 0.8520 (smollab) & 0.6960 \\
    Italian & 0.7139 (7th) & 0.8104 (XplaiNLP) & 0.6941 \\
    Arabic & 0.5771 (4th) & 0.6884 (CEA-LIST) & 0.5133
     \\
    \bottomrule
  \end{tabular}
\end{table}

Our system achieved the best performance for English with a macro F1 score of 0.8052, significantly outperforming the baseline (0.5370). For German, our system ranked third with a macro F1 score of 0.8013, competitive with the top-performing team (0.8520). In Italian, our system achieved a macro F1 score of 0.7139, ranking seventh but still substantially above the baseline (0.6941). For Arabic, our system ranked fourth with a macro F1 score of 0.5771, which is above the baseline (0.5133) but considerably lower than the top-performing team (0.6884).

\subsection{Multilingual Results}

Table \ref{tab:multilingual_results} shows the results of our system in the multilingual setting, where DeBERTa-based models were trained on data from the monolingual setting and evaluated on the multilingual test data.

\begin{table}
  \caption{Multilingual Results (Macro F1 scores)}
  \label{tab:multilingual_results}
  \begin{tabular}{lccc}
    \toprule
    Setting & QU-NLP (Our Team) & Top Team & Baseline \\
    \midrule
    Multilingual & 0.6692 (8th) & 0.7550 (TIFIN India) & 0.6390 \\
    \bottomrule
  \end{tabular}
\end{table}

In the multilingual setting, our system achieved a macro F1 score of 0.6692, ranking eighth among all participating teams. While this performance is above the baseline (0.6390), it is notably lower than our monolingual results for English and German. This suggests that the multilingual model faces challenges in effectively learning shared representations across languages, possibly due to linguistic differences or imbalances in the training data. A clear limitation in our used DeBERTa-based model is coming from the inclusion of sentences in Arabic as part of the testing dataset, whereas the model was not trained on the Arabic language as part of the cross-lingual sequence training explained earlier. 

\subsection{Zero-shot Results}

Table \ref{tab:zeroshot_results} presents the results of our system in the zero-shot setting, where models were evaluated on languages not seen during training.

\begin{table}
  \caption{Zero-shot Results (Macro F1 scores)}
  \label{tab:zeroshot_results}
  \begin{tabular}{lccc}
    \toprule
    Language & QU-NLP (Our Team) & Top Team & Baseline \\
    \midrule
    Romanian & 0.8126 (1st) & 0.8126 (QU-NLP) & 0.6461 \\
    Polish & 0.5165 (13th) & 0.6922 (CEA-LIST) & 0.5719 \\
    Ukrainian & 0.6168 (8th) & 0.6424 (CSECU-Learners) & 0.6296 \\
    Greek & 0.4057 (11th) & 0.5067 (AI Wizards) & 0.4159 \\
    \bottomrule
  \end{tabular}
\end{table}

Our system demonstrated varying performance across the zero-shot languages. For Romanian, we achieved the best performance among all teams with a macro-F1 score of (0.8126), significantly outperforming the baseline (0.6461). This suggests that our sequential fine-tuning approach effectively transferred knowledge to Romanian, possibly due to linguistic similarities with the training languages.

However, for Polish, Ukrainian, and Greek, our system's performance was less impressive. In Polish, we ranked 13th with a macro-F1 score of (0.5165), which is below the baseline (0.5719). In Ukrainian, we ranked 8th with a score of 0.6168, slightly below the baseline (0.6296). In Greek, we ranked 11th with a score of 0.4057, slightly below the baseline (0.4159).

\section{Discussion}

Our participation in the CheckThat! 2025 Task 1 on subjectivity detection yielded several insights into the effectiveness of different approaches for this task across languages and evaluation settings. In this section, we discuss our findings, analyze the strengths and limitations of our approach, and explore potential avenues for improvement.

\subsection{Analysis of Model Performance}

The performance of our systems varied considerably across languages and evaluation settings, revealing several interesting patterns:

\begin{itemize}
    \item \textbf{Strong Monolingual Performance}: Our models performed particularly well in the monolingual setting for English and German, achieving F1 scores of 0.8052 and 0.8013, respectively. This suggests that our feature-augmented transformer architecture effectively captures markers of subjectivity in these languages.
    
    \item \textbf{Varying Cross-lingual Transfer}: The effectiveness of cross-lingual learning transfer varied significantly across target languages. The outstanding performance on Romanian (F1=0.8126) in the zero-shot setting demonstrates that our sequential fine-tuning approach can successfully transfer knowledge to linguistically similar languages. However, the relatively poor performance on Polish, Ukrainian, and Greek suggests limitations in transferring to more distant languages.
    
    \item \textbf{Multilingual vs. Monolingual Trade-off}: Our multilingual model (F1=0.6692) underperformed compared to our best monolingual models, highlighting the challenges of developing a single model that performs well across multiple languages simultaneously. In addition, the Arabic language was not included as part of the cross-lingual sequence training of the model evaluated using the multilingual dataset. 
\end{itemize}

\subsection{Feature Contribution Analysis}

To understand the contribution of different features to our DeBERTa-based model's performance, we conducted an ablation study on the English, German and Italian monolingual models. Table \ref{tab:ablation_study} presents the results of this analysis. Trainings of the monolingual languages forllowed the same sequence of languages [German → Italian → English] as performed in the full models' trainings.

\begin{table}
  \caption{Ablation Study on German, Italian, and English, monolingual models with the sequence of [German → Italian → English] during training}
  \label{tab:ablation_study}
  \begin{tabular}{lccc}
    \toprule
    Model Configuration &  Macro F1 (German) & Macro F1 (Italian)& Macro F1 (English)\\
    \midrule
    Full Model (DeBERTa + TF-IDF + Gating)  & 0.8013 & 0.7139 & 0.8052\\
    DeBERTa Only  & 0.5866 & 0.7040& 0.7974\\
    DeBERTa + TF-IDF (No Gating)  & 0.5245& 0.7234 & 0.7707\\
    Full Model (without cross-lingual training)  & 0.8013 &0.6920 & 0.7818 \\
    \bottomrule
  \end{tabular}
\end{table}

The ablation study reveals that each component of our model contributes to its overall performance:

\begin{itemize}
    \item The base DeBERTa model alone achieved a respectable macro-F1 score of (0.5866, 0.7974, 0.7040) for German, Italian, and English consequently, demonstrating the strong foundation provided by the pre-trained language model when combined with with cross-lingual sequence training.
    
    \item Adding TF-IDF features without the gating mechanism improved performance to (0.7234) for Italian language only, indicating that lexical features do not provide complementary information to the contextual embeddings for all languages.
    
    \item The gating mechanism further improved performance, allowing the model to dynamically balance the contribution of lexical features complementing information gained from the contextual embeddings for English and German languages. 
    
    \item The full model, combining DeBERTa, TF-IDF features, and the gating mechanism, achieved the best performance of (0.8052, 0.8013) for English and German consequently, confirming the value of our feature-augmented approach.
    \item The cross-lingual sequence training positively enhanced the monolingual models' results, where training the full monolingual DeBERTa-based model without the cross-lingual sequence training achieved lower results of (0.7818, 0.6920) compared to (0.8052, 0.7139) for English and Italian languages consequently.
\end{itemize}

\subsection{Language Order in Cross-lingual Training }
\begin{table}
  \caption{F1-macro results for the English, German and Italian Monolingual Models based on language order during training}
  \label{tab:ablation_study}
  \begin{tabular}{lccc}
    \toprule
    Language Order &  Macro F1 (German) & Macro F1 (Italian)&  Macro F1 (English) \\
    \midrule
     (German → Italian → English) & 0.8013 & 0.7139 & 0.8052\\
     (English → Italian → German) & 0.8195 & 0.7787 & 0.7818\\
    (German → English → Italian) & 0.8013 & 0.8033 &0.7839 \\
   
    \bottomrule
  \end{tabular}
\end{table}

The results of our cross-lingual subjectivity detection experiments demonstrate a notable sensitivity to the ordering of language fine-tuning. The results presented in Table~\ref{tab:ablation_study} demonstrate that the ordering of languages during cross-lingual fine-tuning has a significant impact on model performance across English, German, and Italian. Notably, the model trained in the sequence [\textit{English → Italian → German}] achieves the highest F1 score on German (0.8195), and a strong improvement on Italian (0.7787), albeit with a slight drop in English performance (0.7818). In contrast, the sequence [\textit{German → Italian → English}] results in the lowest Italian score (0.7139), while preserving high performance on German (0.8013) and English (0.8052). Interestingly, training in the order [\textit{German → English → Italian}] yields the best Italian performance (0.8033), suggesting a complex interaction between intermediate representations and language-specific features.

These results provide several insights into the dynamics of multilingual transfer for subjectivity detection. First, the finding that German performance improves when preceded by English suggests that English provides beneficial representations which transfer well to German, a typologically related language \cite{conneau2020unsupervised, stabler2003structural}. This supports prior work showing that English often acts as a strong base model for multilingual tasks due to its central position in pretrained multilingual language models.

Second, the decline in Italian performance when preceded by German (0.7139) compared to when preceded by English (0.7787) or English and German (0.8033) is indicative of language interference and potential catastrophic forgetting \cite{li2017learning, kirkpatrick2017overcoming}. German’s syntactically rigid and morphologically rich characteristics may interfere with learning semantic cues for Italian, a Romance language that relies more heavily on pragmatic and lexical signals of subjectivity \cite{pires2019multilingual}.

Third, the sequence [\textit{German → English → Italian}] achieving the highest Italian F1 score implies a positive cumulative effect when both typologically diverse languages precede Italian. This ordering may allow the model to retain robust representations for both syntactic (from German) and semantic-pragmatic (from English) subjectivity features before learning Italian, thereby enabling better generalization.

Lastly, the variations in English scores across the setups (ranging from 0.7818 to 0.8052) suggest that English benefits from being either the final or intermediate fine-tuning target but may degrade when trained first—likely due to subsequent overwriting of its learned representations. This aligns with recent findings on cross-lingual anchoring effects, where the initial language in training can disproportionately shape the shared representational space \cite{muennighoff2023crosslingual}.

\section{Conclusion}

In this paper, we presented our approach to the CheckThat! 2025 Task 1 on subjectivity detection, which challenged participants to distinguish between subjective and objective sentences in news articles across multiple languages. Our system leveraged feature-augmented transformer architectures, combining the contextual understanding capabilities of pre-trained language models with statistical and linguistic features specifically designed to capture markers of subjectivity.

Results demonstrated the effectiveness of our approach, particularly in the monolingual setting for English and German, and in the zero-shot setting for Romanian. The strong performance on Romanian highlights the potential of our sequential cross-lingual fine-tuning approach for transferring knowledge to linguistically similar languages. However, the varying performance across languages and evaluation settings also revealed challenges in developing truly language-agnostic models for subjectivity detection. The ablation study confirmed the value of our feature-augmented approach, showing that each component of our model contributed to its overall performance. 

Our findings reinforce the importance of language order in cross-language fine-tuning and suggest linguistic proximity (i.e. how similar two languages are to each other), and task-specific signal transfer (i.e. how well the model can recognize and use these specific opinion-indicating cues from one language when trying to understand opinions in another), should all be considered when designing cross-lingual pipelines for subjectivity detection. For instance, a language might use certain common phrases or word endings to show an opinion, while another might rely more on the speaker's tone or context. 

Future work could explore several promising directions for improving multilingual subjectivity detection. These include developing more sophisticated language-specific features, implementing adversarial training techniques to create more language-agnostic representations, generating synthetic training data to address class imbalance, exploring multi-task learning approaches, and developing ensemble methods that combine the strengths of multiple specialized models.

\bibliography{sample-ceur}
\end{document}